\documentclass[conference]{IEEEtran}
\IEEEoverridecommandlockouts
\usepackage{cite}
\usepackage{amsmath,amssymb,amsfonts}
\usepackage{algorithmic}
\usepackage{graphicx}
\usepackage{textcomp}
\usepackage{xcolor}
\def\BibTeX{{\rm B\kern-.05em{\sc i\kern-.025em b}\kern-.08em
    T\kern-.1667em\lower.7ex\hbox{E}\kern-.125emX}}
    
\usepackage{caption}
\usepackage{subcaption}
\usepackage{todonotes}

\usepackage{hyperref}
\usepackage[inline]{enumitem}

\begin{document}

\title{Design of the Secure Smart Scooter with Automatic Lane Detection }

\author{\IEEEauthorblockN{Hamid Reza Ghaeini}
\IEEEauthorblockA{\textit{CISPA Helmholtz Center for Information Security} \\
Saarbrücken, Germany \\
\href{mailto:ghaeini@ieee.org}{ghaeini@ieee.org}}
\and
\IEEEauthorblockN{Nils Ole Tippenhauer}
\IEEEauthorblockA{\textit{CISPA Helmholtz Center for Information Security} \\
Saarbrücken, Germany \\
tippenhauer@cispa.de}
}

\maketitle

\begin{abstract}
The authors present a prototype of an electric scooter that uses computer vision to determine driving lanes and automatically modify vehicle configuration. Scooters steer on sidewalks, bike paths, or car lanes. The government imposes speed limits on the lanes in which electric scooters drive. Featuring computer vision, electric scooters can determine driving lanes and automatically modify vehicle configuration. However, such systems' safety is still a concern, and research into such vehicle systems' safety and security aspects is proceeding. First, we will present our vehicular setup. Then, 
we will present our implemented scooter with safety constraints. Finally, we present our security prototype for such an electric scooter.
\end{abstract}


\section{Introduction}
The transportation systems are evolving day by day with integrated technologies such as adaptive cruise control, lane detection, positioning, artificial intelligence and low power wireless communication. One of the goals of an Intelligent Transportation System (ITS) is to support different modes of transportation and traffic management in order to provide users with more relevant information and a safer, more coordinated and "smarter" use of the transportation network. 

The scooter itself, has computing devices that are responsible for the control of the vehicle, and enhanced mobility features such as automatic lane detection. A popular means of transportation in urban areas is the electric scooter, which coupled with other public transportation options can improve mobility and reduce the need for personal mobility devices~\cite{button2020economic,laa2020survey,schellong2019promise,jiao2020understanding}. E-scooters, however, need to be revisited due to the legal framework for these vehicles, and the technology used in them must be adapted to the vehicle technology requirements and the legal framework. The direct interaction of the vehicle and pedestrian lane makes it challenging to design and implement such a mobility medium's safety and security requirements. According to the legal frameworks of Germany, electric scooters can only drive at a defined maximum speed in pedestrian, cycling, and car lanes. For example, the scooter is limited to 6 KM/H in the pedestrian lanes.  

Driver assistance systems and autonomous vehicles require lane detection to form the driving settings. One of the primary tasks of autonomous driving is detecting the road or drivable portion of an unstructured road. Only colour cameras can detect the road or drivable portion in such a muddled road. The literature has discussed road and lane detection for more than four decades, and many promising solutions exist for automobiles today. The authors of~\cite{bar2014recent} explored road and lane detection methods. Lanes and roads can be perceived in several ways, but monocular vision is the most common. Although most works do not discuss it directly, they acknowledge it as a viable choice. Our scooter is cost-effective and delivers reliable performance thanks to a similar approach, which incorporates computer vision. By the way, our discussion of computer vision performance will not take place in this paper, and this paper concentrates on the secure design of such a scooter.

A cruise control advanced driver assistance system known as an adaptive cruise controller (ACC) automatically adjusts the vehicle's speed and distance from the vehicle ahead to preserve a reasonable safety span. Control depends on device information from vehicle sensors~\cite{marsden2001towards, xiao2010comprehensive, milanes2013cooperative}. Such systems may use a measuring instrument, an optical device, or a camera device. This allows the vehicle to brake when it detects that it is approaching another vehicle and to accelerate again when traffic permits. Similarly, we determine the vehicle's maximum speed based on the computer vision component. We enforce the top speed based on the lane detected by our prototype from the visual component to the motor controller. We smoothly transition from the current speed to the desired speed by logarithmic increasing or decreasing the speed in a two-second time window~\cite{hamid2022smooth}. This is critical for maintaining smooth acceleration to ensure the safety of the system~\cite{mayhew2019impact, ma2021scooter, yang2020safety, haworth2021comparing}.

In our prior work~\cite{hamid2022smooth}, we have presented the safety requirements of an electric scooter and the implementation of those safety requirements. This paper discusses the system's security requirements to maintain functional, control and safety applications secure. To our collective knowledge, this paper is the first paper that presents the requirements for the secure design of electric scooters. Our design can also be extended to other motor controllers, such as autonomous driving systems. The rest of the paper will present the vehicular setup, security requirements of vehicular devices, and our proposed design of the secure scooter.

\begin{figure*}[h]
 	\centering
 	\includegraphics[width=0.7\linewidth]{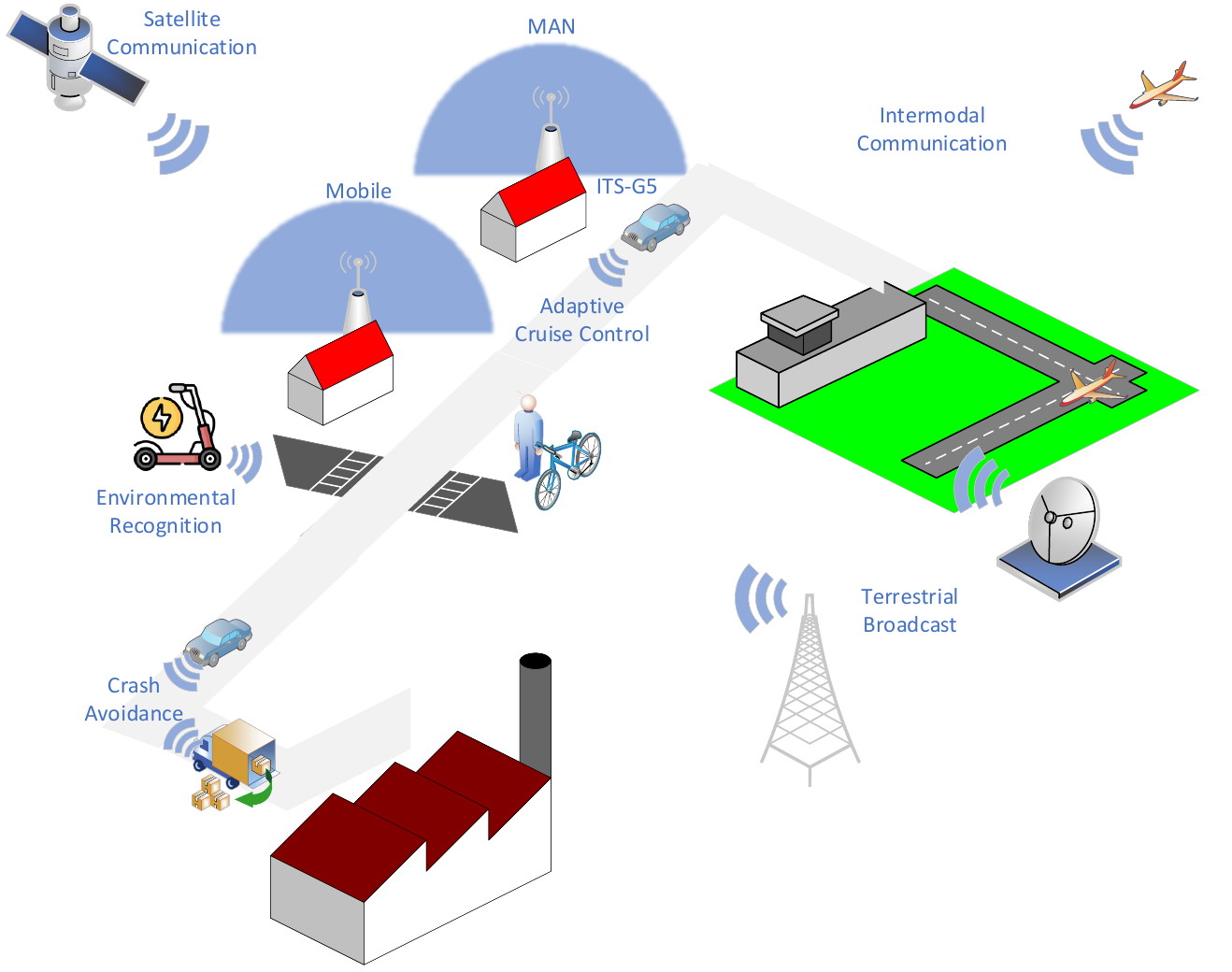}
 	\caption{ITS transportation standards together with our electric scooter.}
 	\label{fig:prototype} 
 \end{figure*}
 
\section{Vehicular Setup}
Motor controllers (MC), Main Control Unit (MCU), sensors, and actuators are the four main components of a scooter. Sensors include, for instance, driver commands, gyroscopes, accelerometers, and so on. The actuators are electric motors, displays, lights, and so on. With the additional camera and Main Control Unit (MCU), lanes can be detected and the maximum speed applied based on the detected lane. 

\begin{figure}[h]
 	\centering
 	\includegraphics[width=\linewidth]{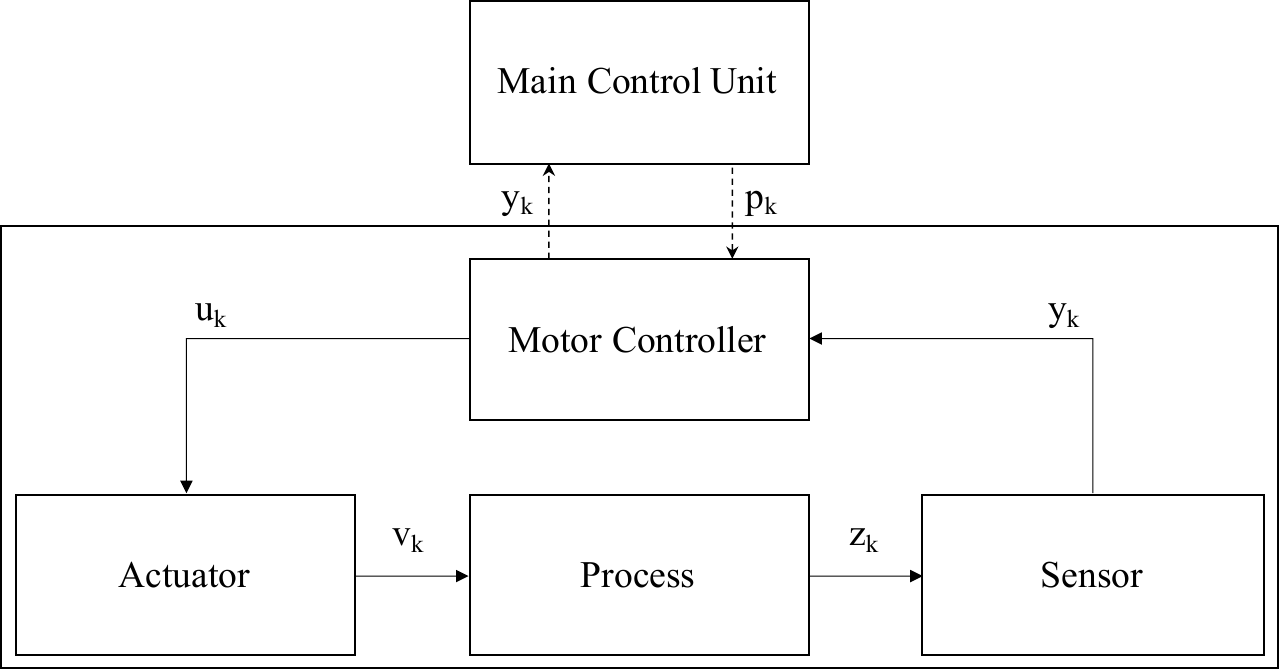}
 	\caption{The control loop of the motor controller. The Main Control Unit receives a copy of sensor readings ($y_k$) and processes them. The high-level control policies will be implied by $p_k$ to the Motor Controller.}
 	\label{fig:controlloop} 
 \end{figure}

Motor controllers (MC), Computer Vision (CV), sensors, and actuators are the four main components of a scooter. Sensors include, for instance, driver commands, gyroscopes, accelerometers, and so on. The actuators are electric motors, displays, lights, and so on. With the additional camera and Main Control Unit (MCU), lanes can be detected, and the maximum speed applied is based on the detected lane. In addition to modelling most control processes, the control loop represents the policies we apply to control the vehicle system.
Figure~\ref{fig:controlloop} shows the motor controller's control loop. Sensors read the state of systems from their processes ($z_k$). Motor controllers receive sensor readings ($Y_k$) from sensors. The motor controller processes sensor readings, the vehicle's space is updated, and actuator commands ($u_k$) are sent to the actuators. Actuators apply actuation commands ($v_k$) to update the system's state.
Motor Controllers send sensor readings to the Main Control Unit (MCU) to measure the system's state. MCUs can perform several advanced state estimations rather than MCs due to the high computing power available there. Lastly, the MC applies the motor controller's policies ($p_k$).

\section{Implemented Scooter}
Here are the major components of our prototype: \begin{enumerate*} \item Main Control Unit (MCU) \item Human-Machine Interface (HMI) \item Motor Controller (MC) \item Environmental Recognition (ER) \item Battery Management System (BMS) \item Cellular Communication \end{enumerate*}.
The main control unit is responsible for processing and decision-making of inputs received from sensors and the MQTT Broker.  
 The MCU applies the required actions to actuators and MCs.  
 The emergency is connected to a camera and is responsible for detecting the current lane and classifying it into three categories: pedestrian, cycle or automobile lanes. 
 It includes a Google Coral Edge TPU that facilitates deep learning and computer vision algorithms. 
 The HMI consists of  consists of 64 $\times$ 64 LEDs, which indicate the current status of the  scooter, with pedestrian traffic,  bicycle or 
 Cellular communication is an NB-IoT connection from Waveshare, including the SIM7080G chip. 
 The figure~\ref{fig:prototype} represents the overall scooter setup, together with other ITS technologies such as satellite communication (GPS), mobile communication (NB-IoT), and environmental recognition.

\begin{figure*}[!ht]
	\centering
	\begin{minipage}{\linewidth}
		\centering
		\subcaptionbox{The maximum possible speed in the pedestrian lane is 6km/h, and also, we requested the maximum gas from the motor controller, the implied policies from MCU maintained the speed at 6km/h.\label{fig:speed-pedestrian}}
		{\includegraphics[width=0.3\linewidth]{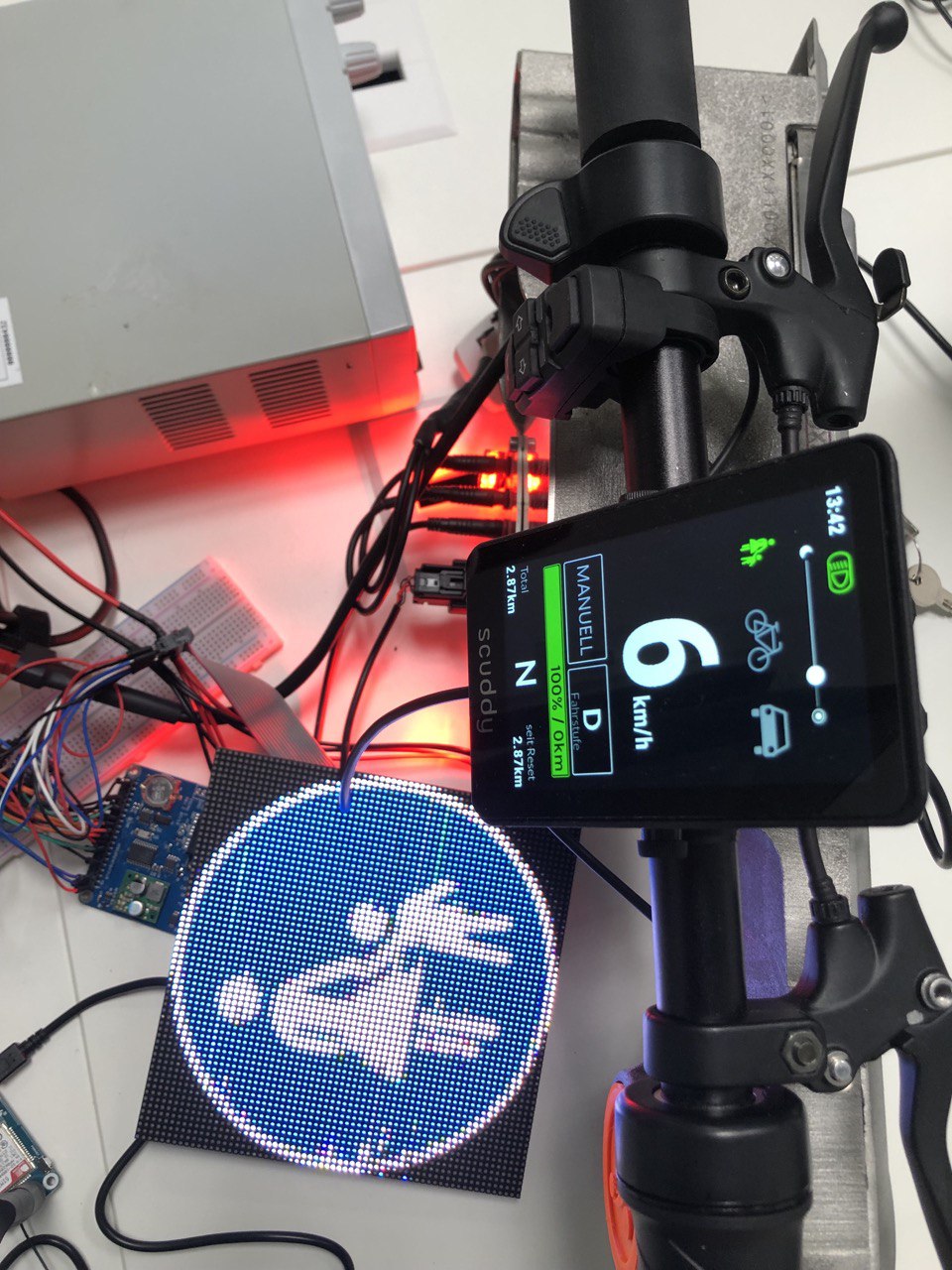}}
		\subcaptionbox{The maximum possible speed in the bike lane is 15km/h, and also, we requested the maximum gas from the motor controller, the implied policies from MCU maintained the speed at 15km/h.\label{fig:speed-bike}}
		{\includegraphics[width=0.3\linewidth]{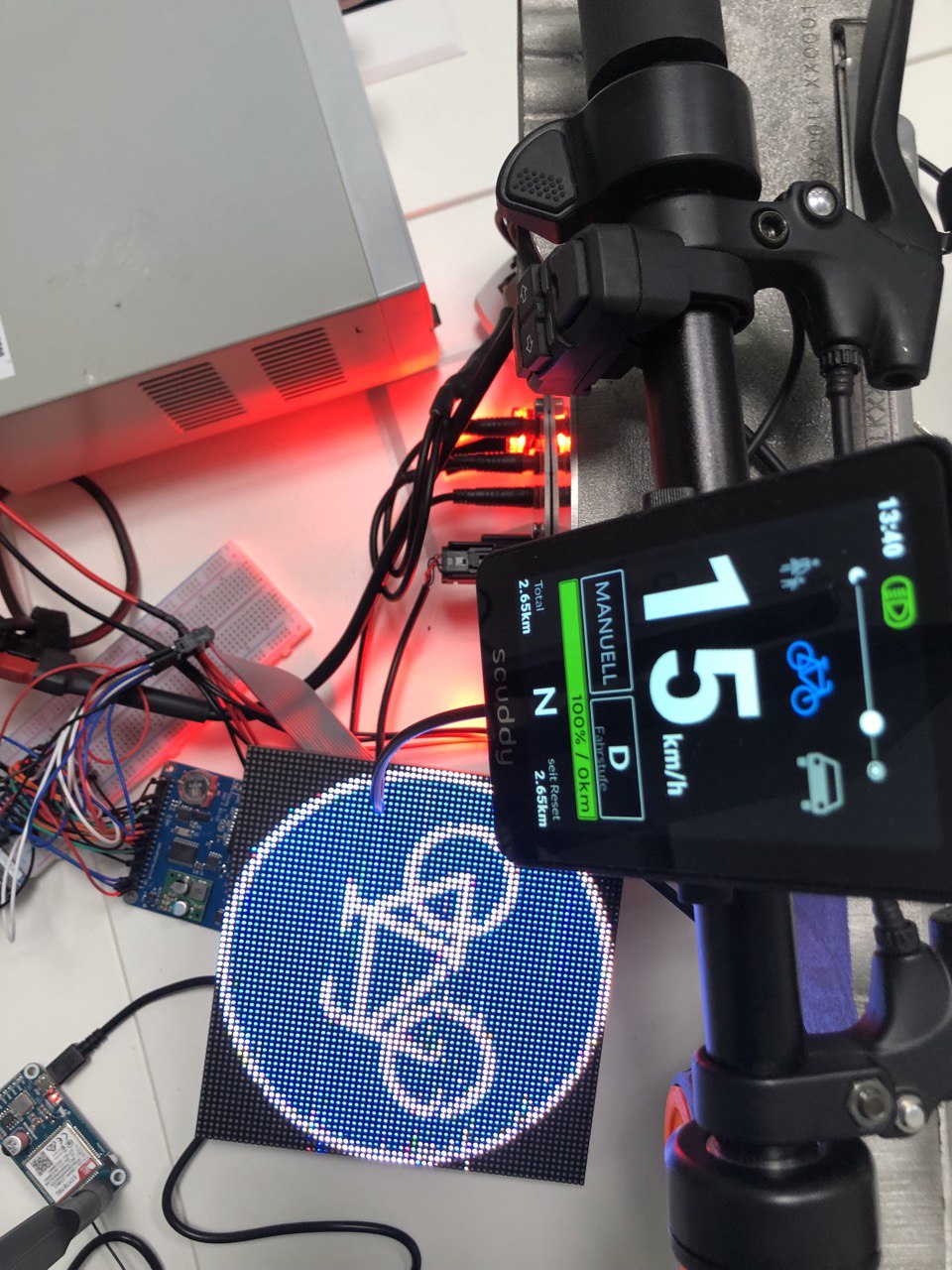}}
  	\subcaptionbox{The maximum possible speed in the car lane is 45km/h, and also, we requested the maximum gas from the motor controller, the implied policies from MCU maintained the speed at 15km/h.\label{fig:speed-car}}
		{\includegraphics[width=0.3\linewidth]{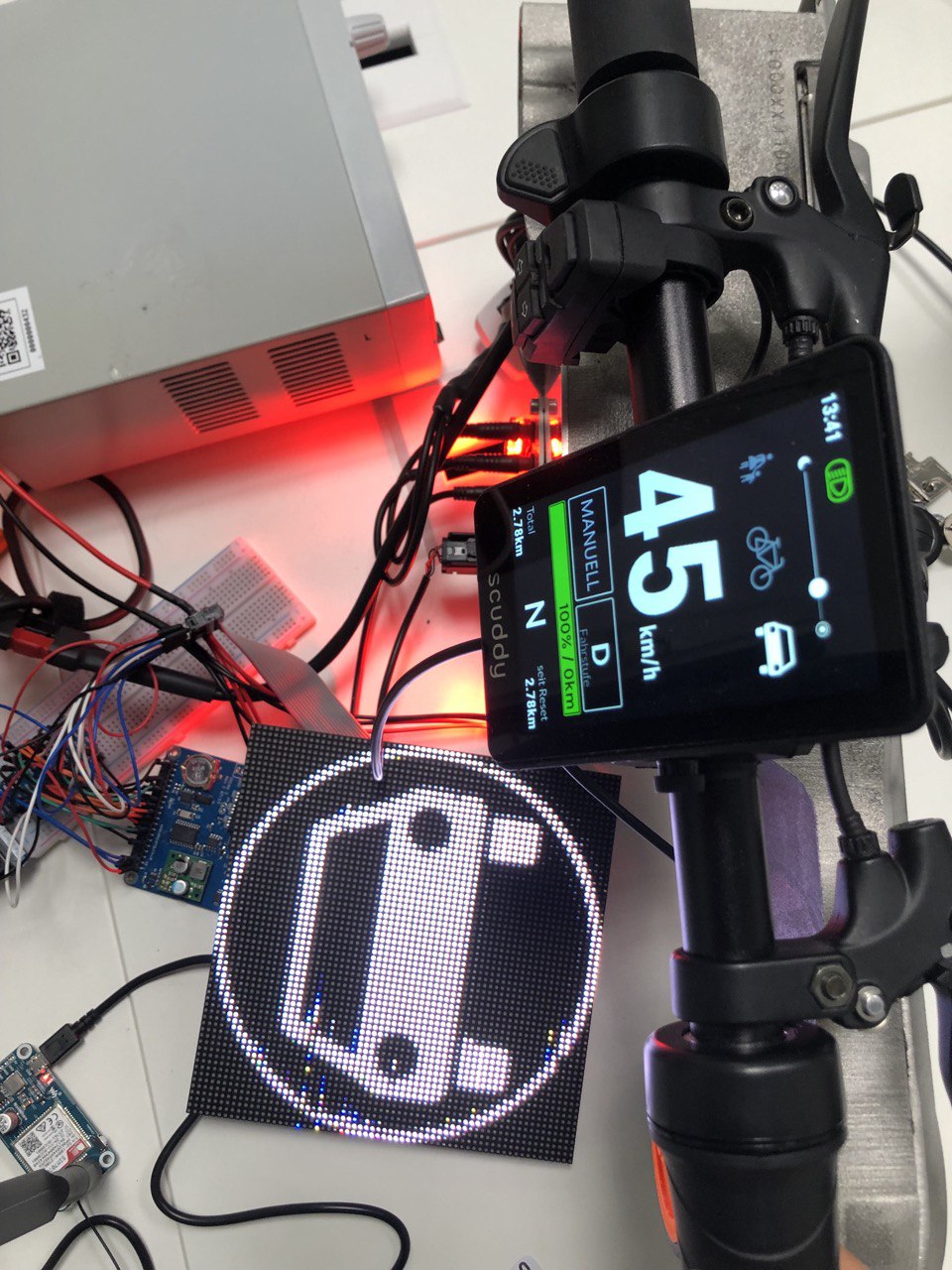}}
		\caption[Deadline]{An illustration of the states of the MCU and implied maximum speed with the request of maximum gas power from the driver.}
		\label{fig:speeds}
	\end{minipage}
\end{figure*}

A CAN bus connects the motor controller to other scooter components. A Raspberry Pi 4 Model B MCU has been attached with a CAN controller, a Coral USB Accelerator, an LED display, and an LED driver. An edge TPU coprocessor, the Google Coral TPU edge enables high-speed machine learning for various systems, mainly for image processing.
Here are a few snapshots of our prototype scooter~\ref{fig:speed-pedestrian}. Designed to enable flexible, efficient, and reliable power supply systems, the Motor Controller is a modified version of VESC. The VESC program provides researchers with resources to help them achieve their goals more quickly. Various engineering challenges arise in VESC environments, ranging from small to large complex vehicle systems. A BLDC motor controller controls the motors directly.

In the same way that brushed DC motor controllers perform a similar function and apply similar styles, BLDC motor controllers also perform a similar job. Their arrangement and execution still differ in some abstract ways. Motor controllers regulate the speed and torque of BLDC motors as well as start, stop, and reverse their rotation. Finally, the scooter is connected to an MQTT broker to send information such as debugging notifications and, when required, positioning pieces of information.

\section{Security Design}
\begin{figure*}[!h]
 	\centering
 	\includegraphics[width=0.7\linewidth]{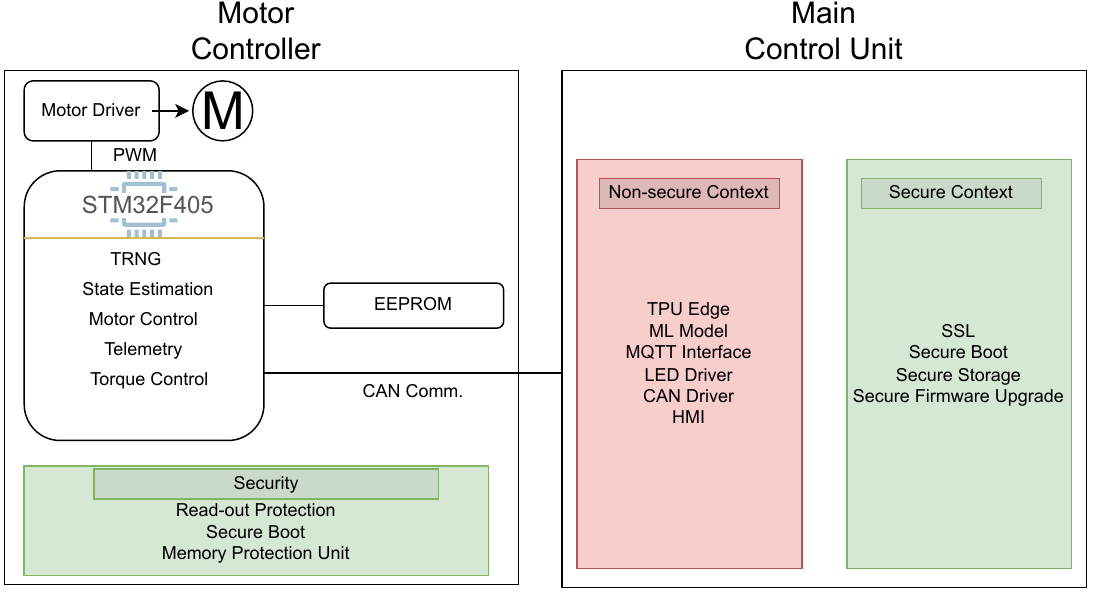}
 	\caption{The overall security design of the electric scooter.}
 	\label{fig:securitydesign} 
 \end{figure*}

Here, we describe the features we included in our security design for the electric scooter. As depicted in Figure~\ref{fig:securitydesign}, these features are read-out protection, secure boot, memory protection unit, SSL protocol, secure storage, and secure firmware upgrade. We will describe each function in this section.

\subsection{Read-out Protection}
Read Out Protection (RDP) is a global read-protection feature for flash memory. The RDP protects the embedded firmware code against copying, reverse engineering, dumping using debug tools or other intrusive attacks. The user should set this protection after loading the binary code into the embedded Flash memory. We used the RDP to protect our secret key that is used during the secure boot.

\subsection{Secure Storage}
Secure Key Storage is available in most general-purpose micro-controller series. This specification orders that it should be feasible to store general-purpose info and key material that assures the confidentiality and integrity of the stored data. The key storage should also provide the atomicity of the operations that modify the storage. Atomicity here means that the entire operation completes successfully or no write is done.

\subsection{Secure Boot}
The Secure Boot (Root of Trust services) is an immutable code, always executed after a system reset. It will check the MCU's static protections, activate MCU run-time protections and then verify the authenticity and integrity of the user application code. This feature will be run before every execution to ensure that invalid or malicious codes will not run. The integrity of the image will be verified against the secret key that is stored in the secure storage.

\subsection{Secure Firmware Update}
The Secure Firmware Update application receives the firmware image via a communication interface and checks its authenticity. It also checks the integrity of the code before lodging it. The firmware update is done on the complete or only on a portion of the firmware image. It can provide single firmware image configuration to maximize firmware image size and dual firmware image configurations to ensure safe image installation. Similar to the secure boot, the integrity of the image will be verified against the secret key that is stored in the secure storage.

\subsection{Memory Protection Unit}
The MPU makes the motor controller more robust and secure by prohibiting the controller applications from corrupting data used by critical tasks (such as the real-time OS).
It also defines the SRAM memory region as a non-executable (execute never XN) to prevent code injection attacks. In addition to those features, MPU can change the memory access attributes. The MPU can protect up to sixteen memory regions, while the MPU is an optional component for memory protection. Including the MPU in the microcontrollers (MCUs) makes them more robust and reliable. 

\subsection{SSL}
MQTT is an IoT communication protocol which is a lightweight publish-subscribe and a machine-to-machine network protocol for Message queue/Message queuing service. It was designed for connections with distant locations that have appliances with resource constraints or restricted network bandwidth.

Transport Layer Security (TLS) and Secure Sockets Layer (SSL) are two main methods to provide a secure communication channel between a client and a server. TLS and SSL are cryptographic protocols which use a handshake instrument to negotiate various parameters and to create a secure link between the client and the server. The MQTT features the SSL protocol to ensure the confidentiality and integrity of the messages in IoT devices.

\subsection{Future Work}
In the future, we will implement our security design and integrate security features into the current scooter controller. We also measure the overall overhead and costs of such a secure design of motor controllers. 
 
\section{Discussion}
We classify the attacker as internal and external attackers. The attacker's target can be the software (firmware), communication between the components, or the physical components. We consider the physical attacks that an attacker can physically damage the device out of the scope of this work. We assume that an attacker gains access to the scooter and tries to manipulate the control units of the scooter.

\subsection{Software Integrity}
The attacker that targets the software integrity can be classified into two types of the attacker. An insider attacker found a vulnerability in the control software and had access to the MCU. The outsider attacker would try to attack the software integrity via communication channels or install a USB-based JTAG (Joint Test Action Group) that communicates with the MCU from outside the running software on the MCU.

\subsubsection{Outsider attacker} The attacker that has access to the MCU from its peripherals would attempt to read the sensitive data and tamper with the control software of the MCU. The attacker's goal is to install malicious firmware or pieces of software and alter the normal cyber-physical behaviour of the scooter, such as speeding the scooter higher than the maximum allowed speed. Our implementation has a secure boot and secure storage function. An outsider attacker attempting to read the sensitive data would fail due to the presence of secure storage that features read-out protection on both the motor controller and the main control unit. Hence, the attacker can not read the secret key stored inside the MCU.

The attacker would try to install malicious software inside the MCU, compromising the software integrity or even uploading new firmware. However, the software integrity will be verified during the secure boot, and the new software or pieces of code will not be loaded during the boot process.

\subsubsection{Insider attacker} The attacker that has full access to the MCU and its memory and the attacker would attempt to read the sensitive data and tamper with the control software of the MCU. The attacker's goal is to install malicious firmware or pieces of software and alter the normal cyber-physical behaviour of the scooter, such as speeding the scooter higher than the maximum allowed speed. The main control unit incorporates the TEE that divides the functions to secure and non-secure zones. The attacker's attempt to read sensitive data stored inside the secure storage would fail due to the isolation concept of the TEE implementation.

The insider attacker with full access to the MCU and its memory can read the sensitive data stored in the motor controller due to the lack of the isolation concept in the MCU. Hence, an attacker that already took access to the MCU, for example, due to a software vulnerability, can read the secret key and generate a correct signature to bypass the secure boot. Nevertheless, such an attack is very complicated and requires a sophisticated strategy of security offensive.

\subsection{Communication Integrity}
The communication mediums of the scooter are internal and external communications. The main control unit is connected to the MQTT broker, and the connection is protected via the SSL protocol. The attacker that might try to alter the communication between the main control unit and the MQTT broker would fail due to the authentication of the client certificate with the SSL protocol or authentication of the client identity with a password.

The internal communication between the scooter devices is via the CAN bus. Due to the lack of message encryption and authentication, the attacker can manipulate internal communications if the attacker gains access to the CAN bus. Nevertheless, we assume the attacker can not attach external devices to the scooter. In any case, CAN communication is not secure, as discussed in the literature.

\section{Conclusion}
We presented our security design of an electric scooter with enhanced safety functionalities, such as detecting the lanes, including pedestrian, bike and car lanes. The scooter consists of two major computing components: the motor controller and the main control unit. The main control unit can use powerful computing resources with advanced security features, such as Trusted Execution Environment. However, the motor controller is usually a real-time controller that lacks advanced security features. We maximized our protection in both components to harden the overall system's security. Thanks to such a security design, we can connect our scooter to the communication networks by using the MQTT protocol and providing ITS features to the current scooters.

\section{Acknowledgment}
This project was funded by the Federal Ministry of Education and Research (BMBF) under the collaborative project of the Kamäleon~\cite{kamaeleon} between HFC Human-Factors-Consult GmbH, Berlin, Constin GmbH, Berlin, Orange-Bike Concept GmbH, Karlsruhe, TÜNKERS, Ratingen, Karlsruhe Institute of Technology, Karlsruhe, Technical University of Braunschweig, Braunschweig, Aachen University of Applied Sciences, Aachen, and CISPA Helmholtz Center for Information Security, Saarbrücken.

\bibliographystyle{abbrv}
\bibliography{ref.bib}

\end{document}